\documentclass{article}

\usepackage[final,nonatbib]{neurips_2020}

\usepackage[utf8]{inputenc} 
\usepackage[T1]{fontenc}
\usepackage{hyperref}
\usepackage{url} 
\usepackage{booktabs} 
\usepackage{amsfonts}
\usepackage{nicefrac}   
\usepackage{microtype} 
\usepackage{booktabs} 
\usepackage{makecell}
\usepackage{microtype}
\usepackage{graphicx}
\usepackage{subfigure}
\usepackage{microtype}
\usepackage{hyperref}
\usepackage{url}
\usepackage{amsthm}
\usepackage{amsfonts}
\usepackage{color}
\usepackage{multicol}
\usepackage{comment}
\usepackage{lipsum}
\usepackage{afterpage}
\usepackage{amsmath}
\usepackage{mathrsfs}
\usepackage{algorithm}
\usepackage[noend]{algorithmic}

\title{Data Shapley Valuation for Efficient Batch Active Learning}

\author{%
  Amirata Ghorbani \\
  Stanford University\\
  \texttt{amiratag@stanford.edu} \\
  \AND
  
  James Zou \\
  Stanford University\\
  \texttt{jamesz@stanford.edu} \\
  \And
  
  Andre Esteva\thanks{Corresponding author.} \\
  Salesforce Research\\
  \texttt{aesteva@salesforce.com} \\
  }

\begin{document}

\maketitle

\begin{abstract}
Annotating the right set of data amongst all available data points is a key challenge in many machine learning applications. 
Batch active learning is a popular approach to address this, in which batches of unlabeled data points are selected for annotation, while an underlying learning algorithm gets subsequently updated.
Increasingly larger batches are particularly appealing in settings where data can be annotated in parallel, and model training is computationally expensive. 
A key challenge here is scale - typical active learning methods rely on diversity techniques, which select a diverse set of data points to annotate, from an unlabeled pool.
In this work, we introduce Active Data Shapley (ADS) -- a filtering layer for batch active learning that significantly increases the efficiency of active learning by pre-selecting, using a \emph{linear time} computation, the highest-value points from an unlabeled dataset. 
Using the notion of the Shapley value of data, our method estimates the value of unlabeled data points with regards to the prediction task at hand. 
We show that ADS is particularly effective when the pool of unlabeled data exhibits real-world caveats: noise, heterogeneity, and domain shift. 
We run experiments demonstrating that when ADS is used to pre-select the highest-ranking portion of an unlabeled dataset, the efficiency of state-of-the-art batch active learning methods increases by an average factor of $6$x, while preserving performance effectiveness.
\end{abstract}

\section{Introduction}

\begin{figure*}[ht]
\centering
\includegraphics[width=\linewidth]{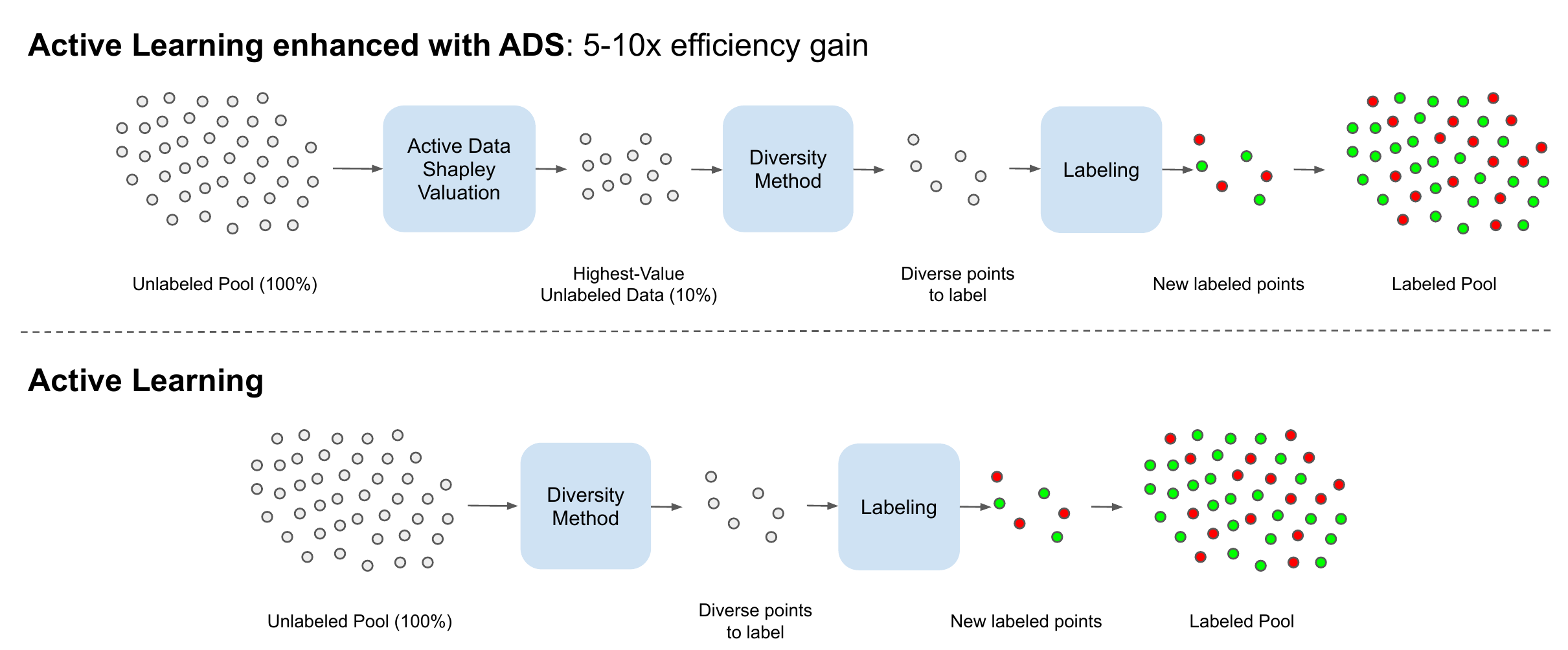}
\caption{\textbf{Effect of Data Shapley Valuation on Active Learning.} (Top) Unlabeled data is filtered via Active Data Shapley Valuation to obtain the small subset that ranks highest by value. This is fed through a standard diversity method for active learning (e.g. CoreSet \cite{sener2017active}) to determine the points to annotate. The significant reduction in data input to the diversity method, coupled with the relatively slow compute times of these methods (around $O(n^3)$) yields a 5-10x improvement in overall computational efficiency. (Bottom) Standard active learning pipeline without shapley valuation.}
\label{fig:graphical_abstract}
\end{figure*}

The last decade has observed remarkable progress in computer vision, as deep learning has pushed the state-of-the-art on nearly every vision task in the field.
%Deep neural networks have shown remarkable performance in various research areas like computer vision, natural language processing and so on. 
The majority of this success can be attributed to supervised learning, typically leveraging a significant number of labeled training examples.
%The majority of these success stories are in the field of supervised deep learning where the deep learning models is trained on a large number of labeled examples. 
Increasing the number of accurately labeled data points tends to improve model performance, as this enables models to better approximate the true distribution of the data. 
Here, the critical drawback is the time and cost of human annotation, particularly in domains requiring massive training datasets (e.g. generalized object classification such as ImageNet~\cite{russakovsky2015imagenet}) or in which the cost of a label is high (e.g. medicine, where expensive expertise is required~\cite{ghorbani2020deep,esteva2017dermatologist}).
%to label; human annotators are asked for their opinion of an input image or text and then their collective opinion is aggregated as the label. 
Techniques that maximize how much models learn from a given training dataset have the potential to improve model performance on the task at hand - a feature which is of particular importance in data-limited and data-expensive regimes.
%It is very desirable to find a way of getting the most performance out of a deep network while spending the least effort. 

The field of active learning focuses on finding \emph{optimal} ways of labeling data. 
This is achieved by coupling a learning algorithm (e.g. a deep model) with an active learner. 
The learning algorithm iteratively trains on a labeled batch of data, then queries the active learner, which selects the next best set of points to be labeled and fed into the learning algorithm. 
This process enables both more efficient learning, and learning equivalently expressive models with smaller labeled datasets, thus reducing the overall cost of annotation.
To date, the field has primarily focused outside of deep learning, on models with lower expressivity and a lower computational training cost~\cite{settles2009active}. 
%There has been a long line of literature focused on active learning but most of the methods were developed and tested for models other than deep neural networks with lower model expressivity and lower computational cost of training~\cite{expressivity}. 
Most of these methods involve labeling a single example at a time, then retraining the model entirely from scratch - a method which is intractable for large-scale convolutional neural networks (CNNs) and other deep models.
%For instance, the majority of existing methods include labelling a single example at a time and retraining the model from scratch which is not affordable in the case of a deep model. 
As a result, recent advances in active learning have been in the \emph{large-batch} setting, such that at each iteration of the algorithm, a large batch of new examples are labeled~\cite{ren2020survey}.

%All existing deep batch active learning methods work through a similar process: First, they compute a heuristic measure of \emph{usefulness score} for each potential unlabeled data points. 
Large-batch active learning methods work through the following process. 
First, they compute a heuristic score of the \emph{utility} of each potential unlabeled data point.
Then, they rank the data points based on this score and choose the next batch for labelling. 
This score is either information-based--how much new information the new point will add--or representation-based---what portion of the data distribution the point represents. 
In choosing the next batch, the utility scores are leveraged to choose a batch which is overall high-ranking, while being \emph{diverse}, such that meaningful signal can be extracted from this data by the model.
Simply selecting the top $N$ ranking points is typically not an effective strategy, as the highest ranking points are likely to be very similar.
%Finally, they seek to solve the problem of \emph{diversity}: choosing the top scored data points might result in choosing data points that are very similar. As a result, instead of choosing $B$ highest scored data points, a more diverse set of $B$ data points is chosen that while being diverse, contains points with high score.

% Both the information- and representation-based methods however, are task-independent. 
% As such, they only consider the unlabeled data, and do not directly consider model performance.
% This limitation prevents the active learner from considering the model's progress on the task at hand in its choice of next batch.
% Here, we introduce a novel approach which circumvents this limitation: data valuation. 
Here we demonstrate that data valuation using the Shapley value of data can filter out most of the unlabeled data (e.g. the bottom 90\%) that is fed into diversity methods. The effect is to significantly increase their efficiency, without impacting their performance, thus allowing them to scale more easily to larger datasets.
We build our utility score around a recently introduced notion in machine learning---data shapley~\cite{ghorbani2019data}---which attempts to compute the contributed value of each data point for a given task. 
The core intuition to our approach is that the incremental value of a training data point to a model varies according to the model's performance on the task at hand.
%The idea is that different training data points contribute on different levels to a prediction tasks and by finding the most valuable ones, we are able to have sample efficiency. 
We design a new algorithm based on this notion, named Active Data Shapely (ADS), which directly optimizes batch selection to enhance the model's final performance.
As a result, ADS performs well even when learning from imperfect and noisy data, which is common in practice.

\paragraph{Our contributions} In this paper, we propose using data valuation to improve batch active learning and propose a new algorithm, ADS, which enhances diversity-based active learning techniques to make them more efficient, while maintaining and even improving performance. We provide comprehensive empirical results to demonstrate that ADS is a reliable method in practice. Our results consider two types of experiments. In the first, we compare diversity techniques against their ADS-enhanced versions using standard curated datasets (CIFAR10~\cite{krizhevsky2009learning}, CINIC10~\cite{darlow2018cinic}, Tiny ImageNet), and demonstrate that ADS increases their efficiency by about 6.4x while preserving the quality of the resultant dataset, as judged by downstream model performance. In the second, we consider more realistic active learning scenarios in which the unlabeled data comes from noisy, heterogeneous, and domain-shifted distributions that are not perfectly matched with the labeled data's distribution. In these challenging settings, we show that ADS improves both the efficiency and effectiveness of the underlying diversity method.
% The key technical contribution of ADS, over previous batch active learning techniques, is the incorporation of a pre-selection step that chooses the most valuable data points for the model at hand using the notion of the Shapley value of data.

\section{Related Work}

This work is related to two families of previous efforts: active learning and data valuation.

\paragraph{Data Valuation} For a model trained on a set of points, data valuation refers to estimating each point's contribution to the model's performance. Traditional methods like cook's distance~\cite{cook1977detection} and its more recent approximations like influence functions~\cite{koh2017understanding} define the importance of a point by the drop in the model's performance after removing the point from training set. 
The Shapley value of data~\cite{ghorbani2019data,jia2019towards}- a notion which satisfies the requisite equitability Shapley axioms~\cite{shapley1953value} - defines a more robust measure of a data point's contribution.
%In addition to satisfying the desirable equitability Shapley axioms~\cite{shapley1953value}, the Data Shapley value~\cite{ghorbani2019data} defines a more robust measure of a data point's contribution. 
The definition of distributional Shapley value~\cite{ghorbani2020distributional}, made it theoretically and empirically possible to estimate the value of new points out of the training set by interpolating the value of existing points. The major drawback with the original Data Shapley method was its computational efficiency - it did not scale to large datasets. The method introduced in ~\cite{jia2019efficient} made it possible to efficiently compute the exact Data Shapley values of points in a KNN model, enabling it's use in this work.

\paragraph{Active Learning}
Directly solving the active learning optimization problem (see Equation ~\ref{eq:max}) is computationally infeasible. Existing methods for active learning propose heuristic methods for choosing the best B points to label. The idea is that data points in the unlabeled pool will provide different amounts of performance gain. The first order approach is to assign a utility score to each point and then choose the top B points. There exists two main family of methods for finding this measure of utility. The first family of methods are focused on \emph{informativeness}. That is, how much additional information each new data point can provide. For a classification problem, the output of a model is usually a probability vector over all the classes. The simplest approaches focus on this probability vector for measure of utility. One natural approach is using the entropy as a measure of prediction uncertainty~\cite{shannon1948bell}; if the model is uncertain in its prediction, the data point will provide information not already stored. Another approach is using the margin - the difference in prediction confidence between the predicted class and the second most probable class~\cite{scheffer2001active, wang2011active, chen2013near}. More complex methods train a number of classifiers on the unlabeled data and use the disagreement amongst them to choose the most informative points~\cite{vasisht2014active,melville2004diverse,guyon2011results} (selection by committee). More recent approaches take a Bayesian approach and usually focus on approximating the parameter posterior~\cite{houlsby2011bayesian}. 

The second family of methods use \emph{representativeness}. That is, choosing the subset of points that are representative examples for a large number of data points in the unlabeled pool. One natural approach is based on clustering ~\cite{li2012active,ienco2013clustering}. The idea is that after clustering is performed, the center of the clusters will be good representative examples of the points in the cluster. There has been a line of work focused on combining the two methods~\cite{du2015exploring}.  For the specific case of batch active learning for deep neural networks, recent uncertainty-based methods like BatchBALD~\cite{kirsch2019batchbald} have been proposed which use the mutual information between model parameters and the data points for selection. A different work~\cite{ducoffe2018adversarial}, approximates the distance of each point to the decision boundary (by running adversarial attacks on the model) and selects the points closest to the boundary. Representative selection methods such as Core-set selection~\cite{sener2017active} methods, seek to select the most representative subset of data points. Both families of methods can suffer from the diversity problem in the large batch setting: Selecting the most uncertain points to label can result in choosing similar points close to the decision boundary and selecting the most representative points can result in only sampling similar points from a high density area of the distribution. As a result, batch active learning methods that seek to sample a diverse batch of points have been developed~\cite{zhdanov2019diverse,wei2015submodularity,hoi2006batch}. All of the existing methods, however, use measures like uncertainty as a proxy for estimating how much a point can improve the trained classifier's performance. Unlike existing methods that use informativeness or representativeness as surrogate measures of a point's expected contribution to a trained model, we estimate the Data Shapley value of points. The Data Shapley value of a point, by definition, is a measure of its expected contribution to a classifier's performance.

\section{Methodology}

    \begin{figure}[ht!]
    \centering
    \includegraphics[width=1.05\linewidth]{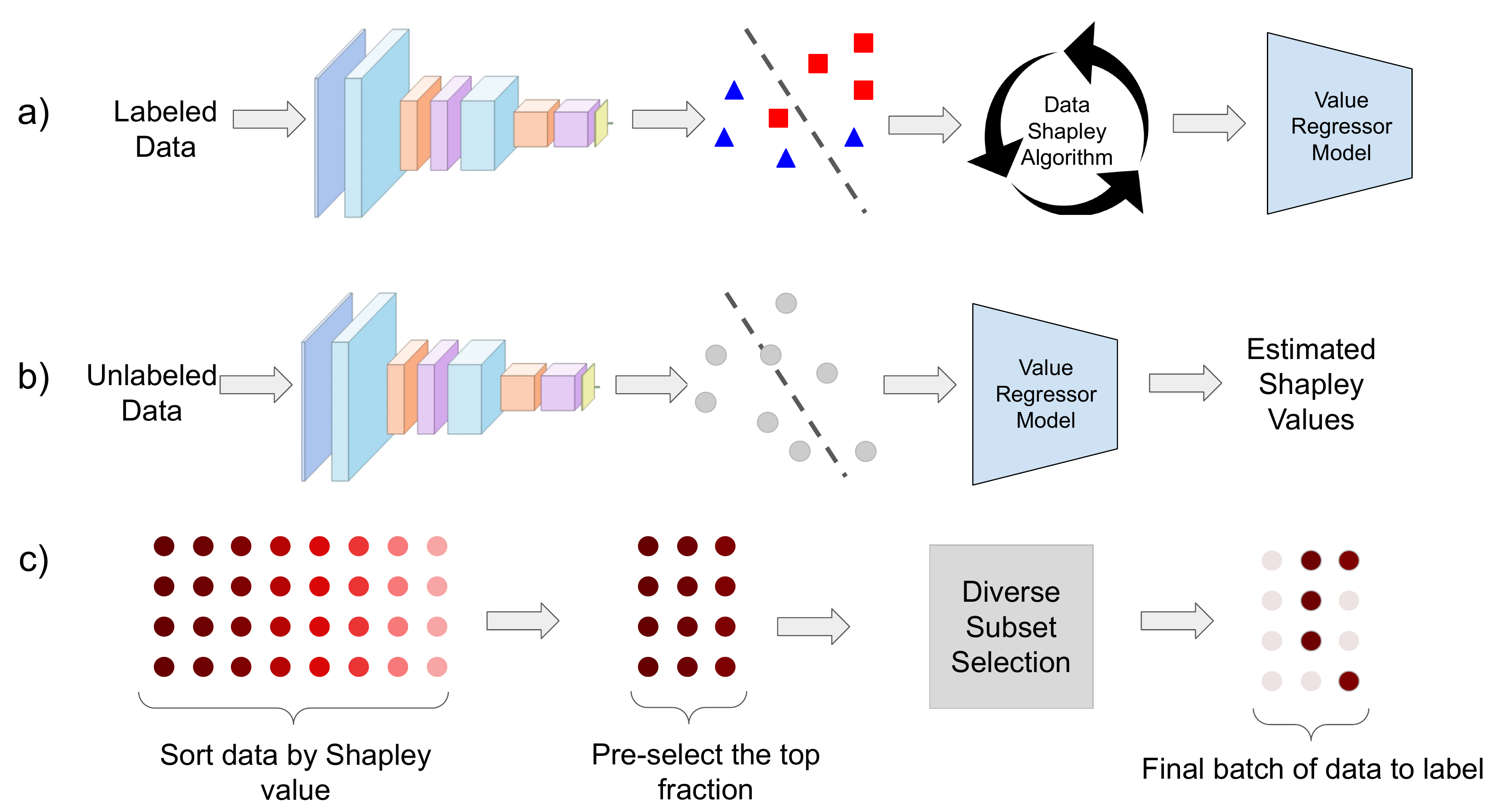}
    \caption{\textbf{Active Data Shapley Enhancing a Diversity Method.} (a) Given a trained model, labeled data is featurized, exact Shapley values are computed, and a regression model is trained to predict Shapley values from features. (b) Unlabeled data is featurized, and Shapley values are estimated with the regressor. (c) Unlabeled data is ranked by the estimated Shapley value. The top fraction is pre-selected and fed into any given diversity method to obtain the final batch of points to label.}
    \label{fig:schematic}
    \end{figure}

\paragraph{Notation} 
We focus on classification problems where the prediction's output space is $\mathcal{Y} = \{1,\dots,C\}$. Let $D$ and $D_T$ annotate the true distribution of training and test data points over $\mathcal{X} \times \mathcal{Y}$, where $\mathcal{X}$ is the space of data points. We are given a set of $i.i.d$ samples from the training data distribution which we refer to as the training set: $\{(x_i, y_i)\}_{i=1}^{n} \sim \mathcal{D}$. For simplicity, we use the data points and their index interchangeably, e.g. $N=\{1,\dots,n\}$ for the training set. In this work, we specifically think about subsets of the training data, that is, $\{(x_i, y_i)\}_{i \in s}$ where $s \subseteq N=\{1,\dots,N\}$. A learning algorithm $\mathcal{A}$ is a black box that takes in a subset of data and outputs a predictive model. In our work, $\mathcal{A}$ is equivalent to training a deep convolutional neural network using stochastic gradient descent. Active learning optimizes over a chosen performance metric $v$ (e.g. 0-1 loss). For simplicity, we use $v(s)$ to refer to the test performance of a model trained on subset $s$, and define $v(s) := \mathbb{E}_{(x, y) \sim \mathcal{D}_T} v(\mathcal{A}(s))$. There exists a labelling function $o$ that returns each example's true label: $o(x) = \mbox{argmax}_{y \in \mathcal{Y}} P_\mathcal{D}(y|x)$. 

\paragraph{Active Learning} 
Improving an existing model's performance requires new labeled training data. Active learning seeks to find and label the smallest number of points to achieve the greatest performance improvement. The active learning setup is as follows: We begin with a given data subset $s^0$, of labeled data points, which we call the ``initial pool''. At each step of the algorithm, we have a labelling budget (the number of times we can query the labelling function) to label $B$ new data points. The first step can be formally written as the following optimization problem:

\begin{align}
\centering
\label{eq:max}
\max_{s^1 \subseteq N \setminus s^0: |b| = B} v(s^0 \cup s^1)
\end{align}

The goal here is to label a size $B$ subset of the unlabeled data that results in the maximum increase in the trained model's performance on the test data. The goal at the $t$'th step is similar. The active learning algorithm uses the previous batches of labeled data to select $B$ new points for labelling from the remaining unlabeled data points. True test performance is approximated using a subset of $M$ data points from the test distribution $\{(x^v_j, y^v_j)\}_{j=1}^{M} \sim \mathcal{D_T}$. From the equation, it's clear that the problem of batch active learning is difficult to solve on two fronts: (1) We have to estimate the the resulting performance without having the labels. (2) The search space is combinatorially large.

\paragraph{Data Valuation}
Data valuation seeks to assign value to individual training datum. Given a model trained on the training dataset and its test performance $v(N)$, valuation seeks to find how much each data point contributed to this performance. In other words, how to divide $v(N)$ among individual datum; for each data point $(x_i, y_i)$, we find to find $\phi_i$ such that $\sum_{i=1}^N \phi_i = v(N)$.  Introduced in ~\cite{ghorbani2019data}, \emph{Data Shapley} provides a solution that satisfies the Shapley equitability axioms:

\begin{itemize}

    \item \textbf{Null element} If $(x_i, y_i)$ results in zero change in performance if added to any subset of $[N] \setminus \{i\}$ , then it should be given zero value. 
    \item \textbf{Symmetry} If two distinct data points  $(x_i, y_i)$ and  $(x_j, y_j)$ results in the same change in performance if added to any subset of $[N] \setminus \{i, j\}$, they should have equal value.
    \item \textbf{Linearity} If the performance metric $v(.)$ is a linear combination of individual metrics, the value should follow the same linearity.
    
\end{itemize}

The Data Shapley value uniquely satisfies the above axioms. For a data point $z = (x, y) \in N$, let $\phi(z)$ be its data Shapley value. We have:

\begin{align}
\label{eq:shapley}
\centering
\phi(z) = \sum_{i = 0} ^ {N - 1} \sum_{s \subseteq N - \{z\}: |s| = i} \frac{v(s \cup \{z\}) - v(s)}{{N - 1 \choose |s|}}
\end{align}

That is, the data shapley value of a datum is a weighted average of its marginal contribution if we add it to all possible subsets of training data. The weight corresponds to the number of subsets that have the same size. It is a measure of the \emph{utility} of each datum. Naturally, a data point with highly positive Shapley value is one that contributes positively to most subsets of the data. A data point with negative Shapley value, on the other hand, is one that on average hurts the performance of the model. The above definition, however, is restricted to the case where the training dataset is fixed. In machine learning, we usually assume the training data is an i.i.d realization of the underlying data distribution.  Assuming the data is coming from the underlying distribution $\mathcal{D}$, one can extend the idea of Data Shapley value to Distributional Shapley value ~\cite{ghorbani2020distributional}. It is proven theoretically and empirically that such a value is Lipschitz continuous for a large group of learning algorithms. More specifically, if $z$ and $z^\prime$ are similar data points (in a proper metric space), their values will be similar. This is very useful in practice; if we have the Shapley values for a subset of data points $s \subseteq N$, we can estimate the value of the remaining data points $N \setminus s$ by interpolating existing values. In practice, once we have ${\phi(x_i, y_i)}_{i \in S}$, we can train a regression model to predict the value of a new data point given its covariates and label. This is important in our setting as we can only compute the data Shapley values for the labeled pool. We can then use these values to train a regression model that predicts an unseen point's value given its covariates. Applying the model to the unlabeled pool, we will have an estimation of the value of each unlabeled point.

\begin{figure}[ht!]
    \centering
    \includegraphics[width=0.8\linewidth]{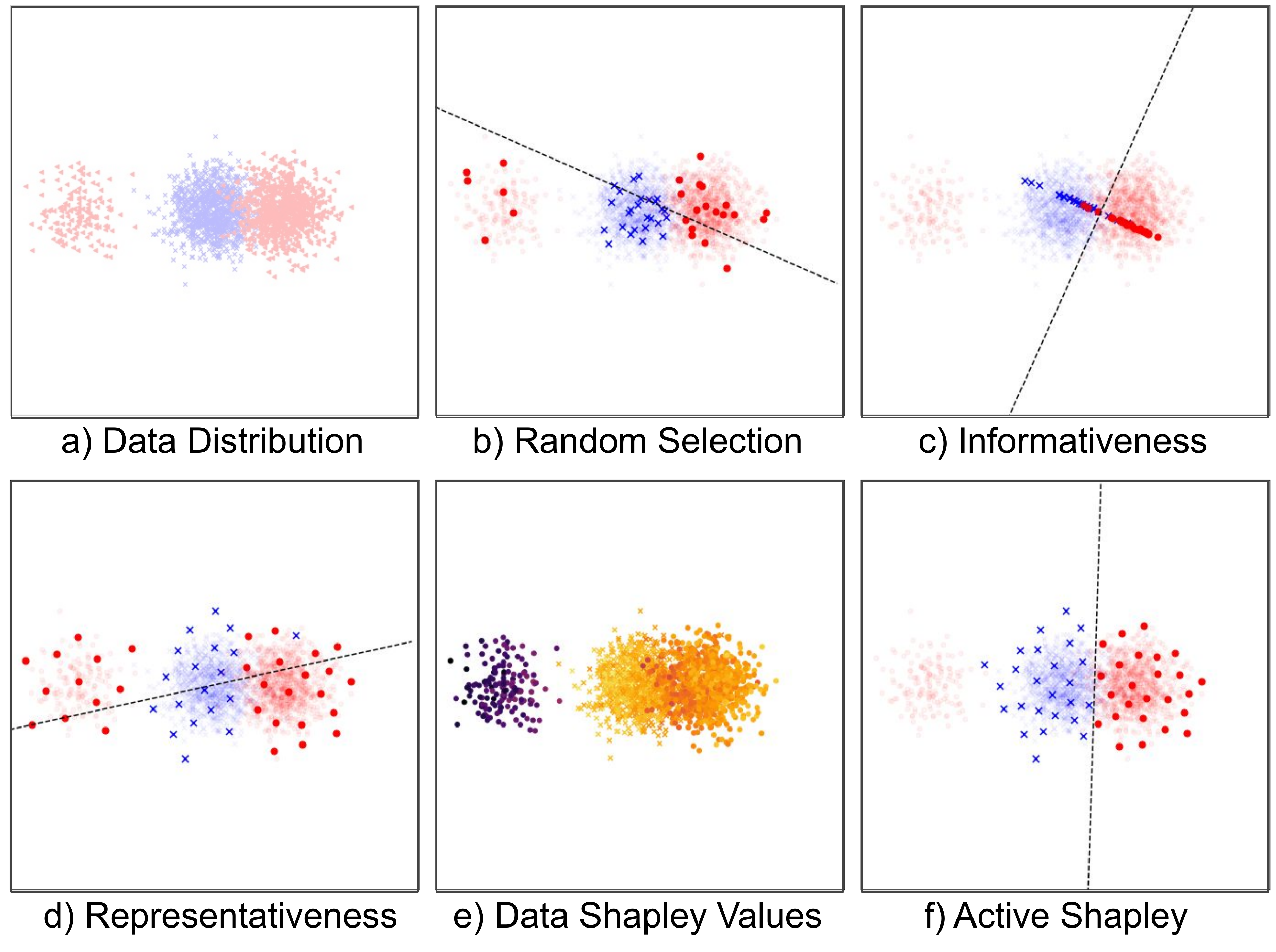}
    \caption{\textbf{Graphical example of various ranking methods.} Suppose we train a binary logistic regression model and its unlabeled data is shown in (a). In (b), (c), (d), and (f) the dotted line represents the decision boundary. (b) Results from choosing a random subset. Darker points indicate those chosen for labeling, lighter points remain unlabeled. (c) Results from choosing the points closest to the boundary, marginally improving performance. (d) Results from choosing a representative set of samples, which does not help final performance. (e) The Data Shapley values. Warmer  color indicates higher Shapley value (e.g.  orange $>$ yellow $>$ purple). Points in the minority cluster get very low values while the other points are useful. (f) Results from removing the points with low Data Shapley value and training on points with high value, which yields an intuitive set of samples to label.
    }
    \label{fig:simple}
    \end{figure}

\paragraph{Batch Active Learning with Shapley Data Valuation}
As mentioned above, batch active learning seeks to find the subset of points that result in the largest improvement in performance. From Eq.~\ref{eq:shapley}, it can be observed that the Data Shapley value of a point is a robust indicator of its improvement to model performance, since it's an exhaustive average of the point's behavior. Assume we are given the true Data Shapley value of all the unlabeled data points. One way of approaching the optimization problem in Eq.~\ref{eq:max} is to reduce the search space by searching over a subset of $N$ that has high value. Fig.~\ref{fig:simple} shows a graphical example of active learning and how this can help. Here, we have a binary classification problem with a logistic regression model in which the class \emph{red} is a mixture of two distributions: a majority group and a minority group (Fig.~\ref{fig:simple}(a)). The points chosen by two simple active learning algorithms, one that works based on uncertainty and another one that works based on representativeness are shown(Fig.~\ref{fig:simple}(c-d)). It can be seen that using either the uncertainty-based or representativeness method will result in poor performance, since performance in both cases is hindered by the minority group in the red class. 
Points in this minority cluster have low data Shapley values because their presence actually hurts the accuracy of a simple linear classifier  (Fig.~\ref{fig:simple}(e)). Removing the low value points from the search space first and then  sampling a representative set of examples greatly improves the performance  (Fig.~\ref{fig:simple}(f)).

\paragraph{Approximating Data Shapley Values}
From Eq.~\ref{eq:shapley} we can see that the exact computation of Data Shapley values is infeasible in most realistic scenarios. One way to approximate them is by the TMC-Shapley algorithm introduced in the original Data Shapley paper ~\cite{ghorbani2019data,jia2019towards}. The idea is that an equivalent form of Eq.~\ref{eq:max} is:
\begin{align}
    \centering
    \label{eq:tmc}
    \phi(z) = \mathbb{E}_{\pi \sim \Pi} [v(s_\pi^z \cup \{i\}) - v(s_\pi^z)]
\end{align}
where $\Pi$ is the set of all permutations of $N$ and $s_\pi^z$ is the set of points that appear before $z$ in permutation $\pi$. Using this form, one can use Monte-Carlo sampling to estimate $\phi(z)$ as follows: sample a permutation of $N$, go over the sampled permutation one point at a time, train a new model on all the observed points, and truncate once adding a new point results in a small marginal change. This results in one Monte-Carlo sample for the value of every data point in $N$. Iterating on this process, you can have an arbitrarily accurate approximation of data Shapley values. 

Although the TMC-Shapley algorithm has been shown to be efficient for simple models~\cite{ghorbani2019data,jia2019towards,ghorbani2020distributional}, for a deep neural network it's infeasible to implement. The reason being that it requires retraining the same model on the order of $O(N^2 \log N)$ times which is impractical in the case of deep neural networks. Luckily, existing work has dealt with this problem~\cite{jia2019empirical} -- using a K-Nearest-Neighbor model, exact Shapley values can be computed efficiently using dynamic programming. The idea can be utilized in deep learning as follows. We focus on the value of points not in their original space (e.g. pixels), but in a learned representation space. We use the pre-logit layer of a trained deep neural network as a good representation space for the classification problem at hand. In this space, training a simple model (like a KNN-classifier) will result in comparable accuracy to that of the original model. We then compute the Data Shapley values using the mentioned method. Note that, by doing so, we ignore the data point's contribution to the representation learning part of the model and only focus on its contribution to the prediction. In this work, we assume a KNN-classifier on top of the learned representation for two reasons: (1) In practice, applying a KNN model on top of a deep network's pre-logit layer achieves similar accuracy to the model's accuracy. (2) We can use the efficient method introduced in ~\cite{jia2019efficient} which computes \emph{exact} Shapley values in linear time.

\paragraph{Active Data Shapley algorithm}
Looking at Eq.~\ref{eq:shapley}, there is an imminent problem: we don't have the label for points other than the ones in the initial labeled pool. As mentioned above, we can use the Lipschitz continuity property of data values. We compute the Data Shapley value of points in the labeled pool and train a regression model to predict the value of points in the unlabeled pool. Fig.~\ref{fig:schematic} visually describes the algorithm. First, we train the deep learning model using the labeled pool, yielding a good representation extractor. We then pass all of the labeled data points through the model to extract their representations, and apply the KNN-Shapley algorithm to compute their exact values. For a problem with $C$ classes, we use $C$ regression models (e.g. KNN regression) that use labeled data points from  each class to predict Shapley values of unlabeled data points.
That is, for each unlabeled point $x^u$, we predict the Shapley value for $(x^u, y_c)$ for each possible choice of $y_c$, yielding $C$ data values. Now the remaining step is to find an aggregate Shapley value for $x^u$. Aggregate approaching include taking the average, or a weighted average using the model's prediction probability for each class. In practice, we find that taking the optimistic approach of using the max value results in the best performance. That is, $\text{Shapley}(x^u) := \max_{c\in C} \text{Shapley}((x^u, y_c))$. When $C$ is large, this step can be computationally expensive. This is solved by limiting the number of possible classes for each unlabeled point. That is, we only consider values of $y_c$ for which the model's confidence is large enough (top 10 in our experiments).

\paragraph{Batch diversity} Choosing the highest scoring points to form a batch can result in a batch of points that are not diverse enough. Consider an adversarial scenario in which each data point is repeated many times in the unlabeled pool. Here, choosing the highest scoring points would result in choosing repetitions of the high value examples. To tackle this issue, one can pre-select a larger set of points based on their predicted Shapley value and then choose a diverse subset from this subset. To this end, we first pre-select a larger number of data points (2-10 times $B$ depending on the size of unlabeled pool) and then use a diversity selection algorithm (e.g. coreset)~\cite{sener2017active} to select a diverse set of $B$ points.

\section{Experiments}
Following previous work~\cite{sener2017active,ash2019deep}, we focus on the task of image classification and consider both the efficiency and performance effectiveness of ADS-enhanced active learning. In all of our experiments, we use a WideResNet model (16-8 for character classification tasks and 24-10 for all other experiments). We assume 5,000 images of the dataset is given as the initial labeled pool, and at each iteration, we label a new batch of 5,000. We first report efficiency results for the first iteration of the active learning problem using several diversity algorithms: (1) Core-set selection~\cite{sener2017active}, using the greedy version of the algorithm for computational efficiency. (2) K-Medians (3) Batch Active Learning by Diverse Gradient Embeddings (BADGE) \cite{ash2019deep}
% 2- The Deep Bayesian Active Learning (BALD) method~\cite{gal2017deep}. A follow-up work, batchBALD, extends this method to the batch setting. However, it's only computationally feasible on small batch sizes. 
% (2) Confidence, which examples with the least confidence of the predicted class are chosen~\cite{wang2014new} 
Due to computational constraints, for the iterative selection process experiments, we focus on three active leanring algorithms: (1) Core-set selection,
(2) Entropy (high-entropy data points are chosen) (3) Random, in which examples are selected randomly.
%FIXIT explain other baselines and why we are not using them

Following ~\cite{gal2017deep}, at each iteration of the active learning acquisition, we retrain the model from scratch as it helps with the performance. In all experiments, a small set of 500 test examples are used as our validation set for early stopping. The same validation set is used for approximating the Data Shapley values. Results are reported on the rest of the test set.

    \begin{figure*}%[ht!]
    \centering
    \includegraphics[width=\linewidth]{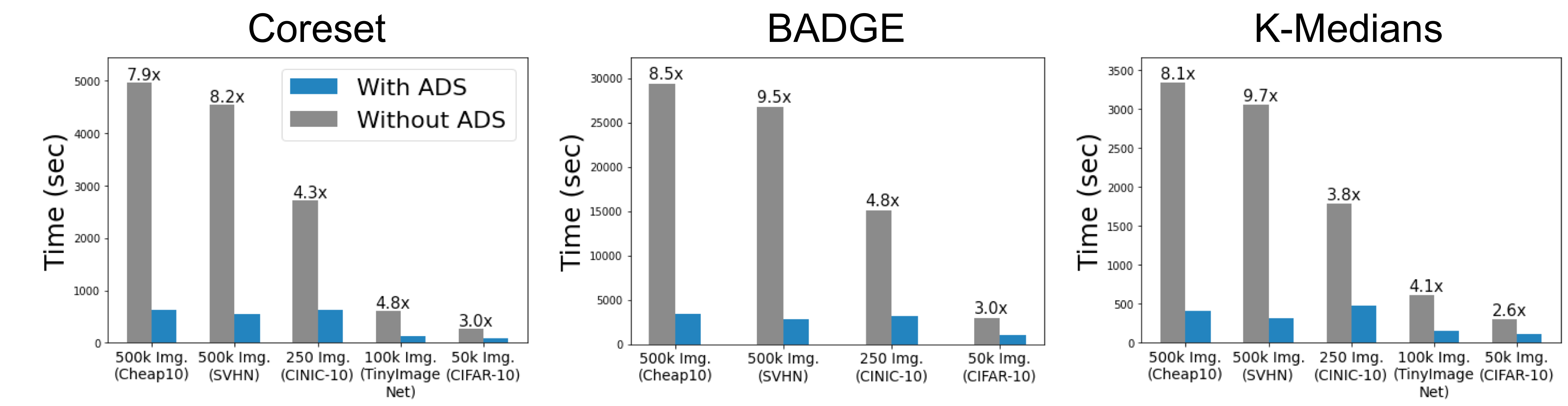}
    \caption{\textbf{The Efficiency Boost of ADS-enhanced Active Learning.} Adding ADS Valuation to three diversity-based active learning pipelines improves the time efficiency. Y-axis shows the average time per batch to compute a diverse set of points to annotate, with and without ADS augmentation. Comparison is shown for Coreset~\cite{sener2017active}, K-Medians~\cite{kmed}, and BADGE~\cite{badge} diversity algorithms. Comparison is shown for the following datasets: Cheap10, SVHN, CINIC10, TinyImagenet, and CIFAR10. The greater the number of images, the greater the overall efficiency boost.}
    \label{fig:time}
    \end{figure*}

\begin{table}[t]
    \caption{\textbf{Performance comparison} Here we compare the performance of the first iteration of active learning algorithm with and without using ADS. We can see that while adding ADS improves the run-time, it preserves the performance. (BADGE is not performed for TinyImageNet as it is computationally prohibitive.) \hspace{10pt}\label{table:compare} \newline}
    \centering
    \begin{tabular}{ l | c | c | c | c| c}
        % \textbf{} & 
        %     \makecell{\textbf{Cheap10} \\ \textbf{500k}} & 
        %     \makecell{\textbf{SVHN} \\ \textbf{500k}} & 
        %     \makecell{\textbf{CINIC-10} \\ \textbf{250k}} &
        %     \makecell{\textbf{Tiny} \\ \textbf{ImageNet} \\ \textbf{100k}} &
        %     \makecell{\textbf{CIFAR10} \\ \textbf{50k}}
        %     \\
        \textbf{} & 
            \makecell{Cheap10 \\ 500k} & 
            \makecell{SVHN \\ 500k} & 
            \makecell{CINIC10 \\ 250k} &
            \makecell{Tiny \\ ImageNet \\ 100k} &
            \makecell{CIFAR10 \\ 50k}
            \\
        \hline
        K-Median & 63.4 & 94.9 & 64.5 & 34.2 & 85.4\\
        K-Medians+ADS & 63.3 & 95.5 & 65.2 & 34.1 & 85.4\\
        \hline
        Coreset & 63.1 & 93.1 & 64.7 & 34.3 & 86.2\\
        Coreset+ADS & 63.4 & 93.4 & 64.7 & 34.4 & 86.4 \\
        \hline
        BADGE & 63.9 & 95 & 70.4 & - & 87.8\\
        BADGE+ADS & 64.3 & 95.4 & 70.4 & - & 87.2
    \end{tabular}
\end{table}    

\subsection{Active Learning Efficiency}
In Fig.~\ref{fig:time}, we compare the time-efficiency of Coreset, BADGE, and K-Medians to their ADS-enhanced versions. These methods scale linearly with the size of the unlabeled pool. ADS-enhanced diversity applies the diversity method on a \emph{pre-selected subset} of high value unlabeled data points, much smaller than the original unlabeled pool. Thus, it can linearly decrease the computational cost. For instance, if ADS removes 90\% of the unlabeled pool in the pre-selection step, then it will be about 10 times faster since the computational cost of KNN-Shapley algorithm is small. The larger the unlabeled pool, the greater the efficiency gain. In all cases considered, ADS-enhancement yields an efficiency gain factor of 2.6-8x. Note that the reported times for ADS include the time it takes to regress and predict the Shapley values. Table~\ref{table:compare} shows that despite the speed-gain, adding ADS to the active learning pipeline can improve the overall performance.

\subsection{Active Learning Effectiveness on Curated Data}
We consider the effect of ADS-enhancement labeling both curated and noisy data. In the curated case, we use standard baseline datasets in which the unlabeled pool of data and the test set both come from the same distribution. All methods on curated data converge to similar performance levels. Here, ADS-enhanced methods either match or slightly outperform their counterparts, as described below and illustrated in Fig.~\ref{fig:effectiveness}(left column).

    % \begin{figure*}[ht!]
    % \centering
    % \includegraphics[width=0.8\linewidth]{Fig1.pdf}
    % \caption{\textbf{Active Learning Effectiveness on Curated Data.} We use a WideResNet model for all experiments. We start with an initial labeled pool of $5,000$ images.  At each step of active learning, a new batch of $5,000$ images are selected and labeled using each method. Coreset is the previous SOTA method which does not filter data by Shapley value. Entropy is another popular algorithm that selects data with the highest prediction entropy to label. The random selection results is also shown.
    % Active Shapley consistently improves or matches the best performing method, while substantially improving the selection speed. See Table ~\ref{table:time} for a computational efficiency comparison with Coreset.}
    % \label{fig:fig1}
    % \end{figure*}

% \paragraph{Datasets} 
We use three benchmark datasets to show variations of the problem. The first is CIFAR-10~\cite{krizhevsky2009learning}, containing 50,000 tiny ($32\times32$) colored images of 10 objects. The second is CINIC-10~\cite{darlow2018cinic}, an interesting variant of CIFAR-10 that contains the same 10 classes as CIFAR-10 but comes from two different sources. 50,000 of its images come from CIFAR-10 images and 200,000 images come from ImageNet(ILSVRC2012)~\cite{russakovsky2015imagenet}, from the same 10 classes. It is chosen in order to better understand how active learning methods work when the unlabeled pool is much larger and the data is more diverse. The third is Tiny ImageNet, chosen to investigate the scenario of having a larger number of classes. Tiny ImageNet has 100,000 images from 200 different object classes. We set the pre-selection threshold according to the size of the unlabeled pool. For CIFAR10, since the pool is small, we pre-select $30\%$ of the points with the highest Shapley value and then apply Coreset at each iteration of batch active learning. For CINIC-10 and Tiny ImageNet, we apply Coreset after pre-selecting $20\%$ of the unlabeled points with the highest value. Note that this could be replicated with any other diversity method.

% \paragraph{Results} 
Experimental results are shown in Fig.~\ref{fig:effectiveness}(left column), where we compare performance across the various methods. We focus on the first five iterations of each algorithm for clarity.
%\andre{Fix this sentence - unclear:} We report the results for complete number of iterations using the area under curve measure in Table~\ref{table:time}. %FIXIT
%For CIFAR-10, we remove the bottom 80\% of the unlabeled data points with lowest value and for the other two experiments we remove 90\%.
It can be observed that in all three datasets, ADS-enhanced coreset essentially matches the performance of Coreset alone, and both techniques outperform the two baselines. This shows the ADS can enhance a diversity method without its data reduction step negatively impacting model performance on curated data.

\subsection{Active Learning Effectiveness on Noisy Data}

    \begin{figure}[ht!]
    \centering
    \includegraphics[width=0.7\linewidth]{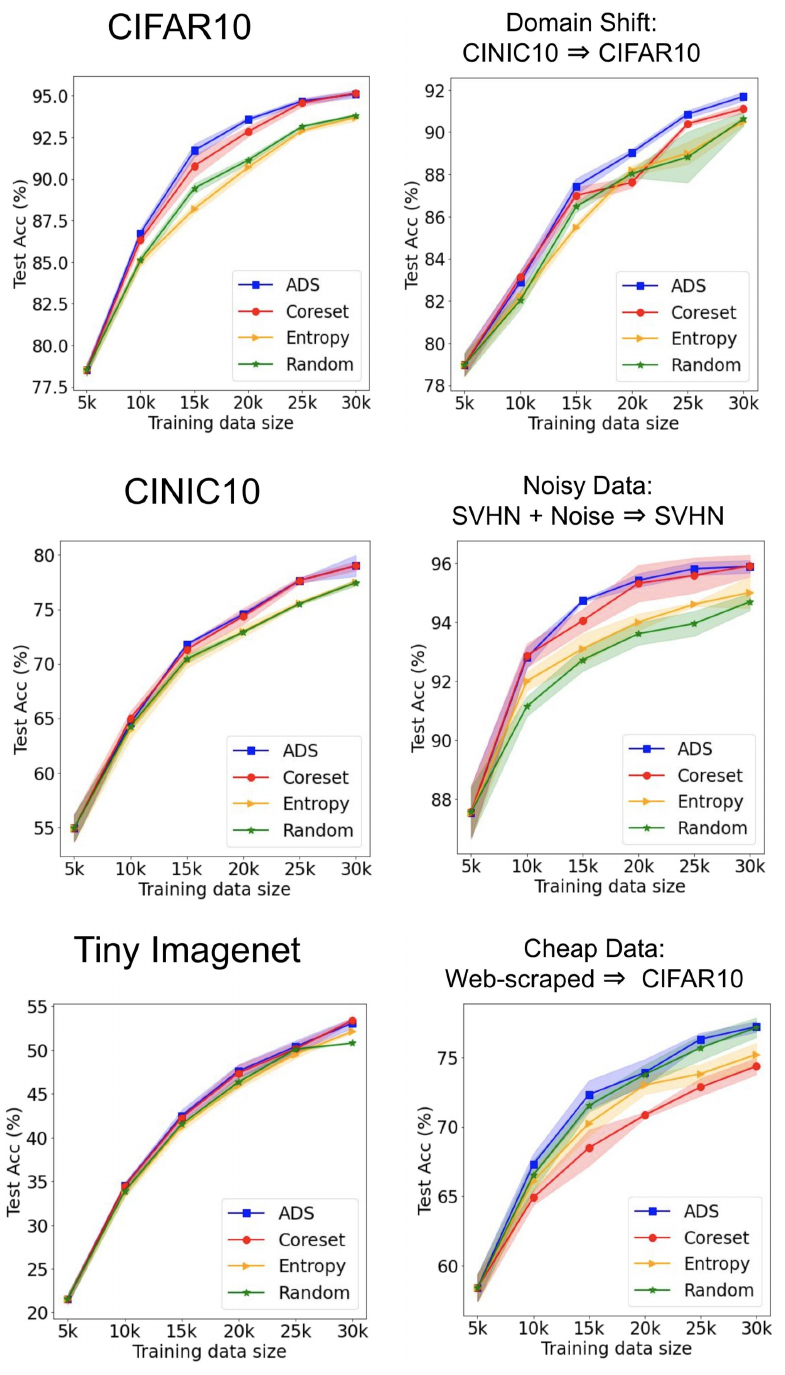}
    \caption{\textbf{ADS-enhanced Active Learning Effectiveness on Curated and Noisy Data.} We use a WideResnet model for all experiments. We compared ADS-enhanced-Coreset to standard Coreset, Entropy, and Random selection. \textbf{Curated Data (Left Column)}. Model comparison on CIFAR10, CINIC10, and TinyImagenet. ADS-enhancement does not negatively impact performance. \textbf{Noisy Data (Right Column)}. We choose three different proxies of real-world scenarios and apply active learning on unlabeled data from one domain and evaluate model performance on the data from another. Domain Shift: Unlabeled data from CINIC10 and performance evaluation on CIFAR10. Noisy Data: Unlabeled data from SVHN with added noise and performance evaluation on SVHN. Cheap, Weakly-labeled Data: Unlabeled data from a fast, low-quality web-scrape, and performance evaluation on CIFAR10. In these three experiments, ADS improves performance primarily by removing low-quality data.}
    \label{fig:effectiveness}
    \end{figure}

The assumption of an unlabeled pool of data points that come from the exact distribution of the labeled pool is unrealistic in real-world setups. Often the unlabeled data is not clean; images can be corrupted by various confounders like noise, distortion, etc. In other cases the unlabeled data comes from various sources, and it is unknown how closely each data source matches the test data distribution. Further, the unlabeled data may be collected quickly, and without expensive curation. We can leverage value from a cheap and low quality unlabeled pool of data points by intelligently selecting its best data points. In the following set of experiments, we simulate such real-world scenarios to evaluate the different learning algorithms.

% \paragraph{Datasets} 
As in the curated data experiments, we use Coreset as the diversity method, and work with both CIFAR-10 and CINIC-10.
Additionally, we work with the street view house numbers (SVHN) dataset, which contains more than 70,000 colored images for the task of digit classification. It contains an extra set of over 500,000 images that we use to mimic the realistic scenario of having a large unlabeled pool.  

Further, we create a web-scraped dataset -- Cheap-10 -- designed to investigate the real-world setting of gathering a large pool of unlabeled data points quickly and inexpensively. To this end, we use the Bing image search engine and search the title of each class in CIFAR-10 with reasonable keywords e.g. ``convertible car'' to scrape the web. We create a data set of 500,000 images (ten times the size of CIFAR-10) in just a few hours of effort. The data set, while containing many valid images, contains lots of out of distribution examples, noisy examples, and mislabeled examples.  Given the large size of the unlabled pool in both datasets, we pre-select only $10\%$ of the unlabeled pool using the estimated Shapley values before choosing a batch of $5000$ using core-set selection.

% \paragraph{Results}
Using the above three datasets, we run three experiments to examine different aspects of noisy data scenarios. The results are shown in Fig.~\ref{fig:effectiveness}(right column). 

The first experiment investigates domain shift. CINIC-10 serves as the unlabeled pool of data, while performance is measured on CIFAR10. CINIC-10 is a mix of CIFAR-10 and ImageNet, thus one can assume that some of the ImageNet images are in-domain while others are not (e.g. a panther image for the cat class). We can see that ADS-enhanced active learning achieves top performance over baselines, since it can find the points that are most helpful in classifying CIFAR-10 images. 

The second experiment models the case in which the given unlabeled data is partially corrupted. We use SVHN's extra training set as the unlabeled pool and corrupt $80\%$ of the images with white noise. The power of the noise is randomly sampled from a Beta distribution to simulate the real-world scenario of having images of varied quality. All active learning methods are better than random selection, and ADS-enhanced active learning achieved the best performance. 

Our final experiment represents a common real-world scenario: gathering curated data is expensive and time-consuming, whereas low-quality unlabeled data is abundantly available. The Cheap-10 dataset contains CIFAR-10 classes, but is corrupted by a large number of out of task and out of distribution images (e.g. a car brand logo for the car class). Here, ADS-enhancement significantly outperforms the other active learning methods, though random selection also performed surprisingly well. 

This set of experiments show that ADS provides a degree of robustness to real-world distributional shifts during active learning -- matching and even enhancing the performance of diversity methods. Considering both the curated data and noisy data sets of experiments, we conclude that using the Shapley Value of data as a metric helps to find the best images for the intended task at hand, while significantly boosting the efficiency of active learning.

\section{Discussion}

We propose Active Data Shapley as a new enhancement step for active learning. 
This is of particular value in the context of deep learning, which needs very large datasets.
Historically, active learning methods have chosen the next best data point, from an unlabeled pool, to be labeled. 
In the context of deep learning, batch active learning is employed as a more scalable method, in which the next best batch of data is selected from an unlabeled pool.
Standard batch active learning methods assign a utility score to unlabeled data points and label a group of highly scored points. 
Unlike prior methods, ADS takes a new approach for computing the utility scores that is built on the notion of the Shapley value of data. 
Data Shapley value is a direct measure of how much a point will help the prediction accuracy of an existing model if that point joins the existing training data points. 
Through extensive experimentation, we show that ADS enhances existing active learning techniques by making them more efficient and more robust to noisy unlabeled data.
We perform experiments that model real-world scenarios in which unlabeled data points are intentionally misaligned with the test distribution, and show that our method significantly enhances existing techniques. 
Of note, ADS is a general framework which can be extended to any other learning model, and future research directions may apply ADS beyond deep learning.
Additionally, we note that the ADS pipeline depends on the quality of the value estimation; if we are unable to properly estimate unlabeled points' values, the method's performance will suffer.

\clearpage
\bibliography{references}
\bibliographystyle{aaai}

\newpage

\appendix

\end{document}